\documentclass{edm_article}
\usepackage{subcaption}
\usepackage{graphicx}
\usepackage[table]{xcolor}
\usepackage{array}
\usepackage{booktabs} 
\usepackage{tabularx} 
\usepackage{ragged2e} 
\usepackage{xspace}
\usepackage{enumitem}

\definecolor{lightgray}{gray}{0.9}
\newcolumntype{P}[1]{>{\centering\arraybackslash}p{#1}}
\newcolumntype{M}[1]{>{\centering\arraybackslash}m{#1}}
\newcommand{\sa}{\textit{SA-RL}\xspace}
\newcommand{\da}{\textit{DA-RL}\xspace}
\newcommand{\saf}{\textit{Suggestion-Assisted RL}\xspace}
\newcommand{\daf}{\textit{Decision-Assisted RL}\xspace}
\newcommand{\llm}{\textit{LLM-based}\xspace}
\newcommand{\rl}{\textit{RL-based}\xspace}
\newcommand{\arl}{\textit{LLM-assisted RL}\xspace}
\newcommand{\perf}{\textit{Post-test Performance Score}\xspace}
\newcommand{\traj}{\textit{Trajectory Quality Score}\xspace}
\newcommand{\comb}{\textit{Combined Score}\xspace}
\newcommand{\pst}{\texttt{PharmaSimText}\xspace}

\usepackage[numbers,sort&compress]{natbib}
\usepackage{balance}

\begin{document}

\title{Towards Generalizable Agents in Text-Based Educational Environments: A Study of Integrating RL with LLMs}

\numberofauthors{3}
\author{
\alignauthor
Bahar Radmehr\\
       \affaddr{EPFL}\\
       \email{bahar.radmehr@epfl.ch}
       \alignauthor
       Adish Singla\\
       \affaddr{MPI-SWS}\\
       \email{adishs@mpi-sws.org}
       \alignauthor
       Tanja Käser\\
       \affaddr{EPFL}\\
       \email{tanja.kaeser@epfl.ch}
}

\maketitle

\begin{abstract}
There has been a growing interest in developing learner models to enhance learning and teaching experiences in educational environments. However, existing works have primarily focused on structured environments relying on meticulously crafted representations of tasks, thereby limiting the agent's ability to generalize skills across tasks. In this paper, we aim to enhance the generalization capabilities of agents in open-ended text-based learning environments by integrating Reinforcement Learning (RL) with Large Language Models (LLMs). We investigate three types of agents: (i) RL-based agents that utilize natural language for state and action representations to find the best interaction strategy, (ii) LLM-based agents that leverage the model's general knowledge and reasoning through prompting, and (iii) hybrid LLM-assisted RL agents that combine these two strategies to improve agents' performance and generalization. To support the development and evaluation of these agents, we introduce \pst, a novel benchmark derived from the PharmaSim virtual pharmacy environment designed for practicing diagnostic conversations. Our results show that RL-based agents excel in task completion but lack in asking quality diagnostic questions. In contrast, LLM-based agents perform better in asking diagnostic questions but fall short of completing the task. Finally, hybrid LLM-assisted RL agents enable us to overcome these limitations, highlighting the potential of combining RL and LLMs to develop high-performing agents for open-ended learning environments.
\end{abstract}

\keywords{Reinforcement Learning, Large Language Models, Text-Based Educational Environments, Learner Models} 

\section{Introduction}
Learner models are foundational to the advancement of educational technologies, serving as a versatile tool for a multitude of applications that enhance both teaching and learning experiences~\cite{Kser2023}. By simulating the interactions and data of students, these computational models provide a safe and controlled environment for teacher training, allowing educators to refine their methods without direct implications on actual students~\cite{DBLP:conf/sigcse/RobinsonJR18}. They also facilitate the development and evaluation of adaptive learning systems~\cite{DBLP:conf/its/DickisonRNHTMH10} or new algorithms~\cite{DBLP:conf/edm/NazaretskyHA19}. Furthermore, they have been applied for testing theories of learning~\cite{DBLP:conf/edm/MacLellanHPK16} and foster collaboration skills in students through interacting with virtual peers~\cite{DBLP:journals/aiedu/Pareto14}.
 
Reinforcement learning (RL) offers a promising avenue for developing these learner models/agents~\cite{DBLP:journals/corr/abs-2107-08828}. Existing works on RL for educational domains have primarily focused on developing techniques for curriculum optimization~\cite{DBLP:journals/tlt/WhitehillM18,pickthemom,zhou_hierarchical_2019,DBLP:journals/cogsci/RaffertyBGS16}, providing tailored hints and feedback~\cite{DBLP:conf/edm/EfremovGS20,DBLP:conf/its/BarnesS08}, and generating educational content~\cite{DBLP:conf/nips/AhmedCEFGRS20,padurean2024neural}.
Only a limited number of works have explored the use of RL-based learner agents that effectively operate in the learning environments~\cite{maclellan_learning_2021,DBLP:conf/iclr/BunelHDSK18}.
However, these RL-based learner agents have been studied for structured tasks with well-defined rules, such as mathematics and logic puzzles. In such environments, RL's capabilities are naturally exploited due to the straightforward definition of state and action representations using engineered features obtained from the existing structure~\cite{DBLP:journals/corr/abs-2107-08828,align,maclellan_learning_2021}. 
However, the reliance on hand-crafted features and engineered state representations limits the ability of these RL agents to be used in unstructured domains and to generalize their learned skills and knowledge across different tasks.
 
Recent advances in generative AI, in particular Large Language Models (LLMs), provide new opportunities to drastically improve state-of-the-art educational technology~\cite{DBLP:journals/corr/abs-2402-01580}. LLMs are capable of generating coherent and contextually relevant content, engaging in meaningful dialogues, and executing specific linguistic tasks without explicit training~\cite{NEURIPS2020_1457c0d6,DBLP:journals/corr/abs-2303-12712}. So far, in education, LLMs have mainly been \textit{applied} for generating educational content~\cite{Kumar2023,DBLP:conf/icer/SarsaDH022,DBLP:conf/icer/PhungPCGKMSS22}, automating grading and feedback processes~\cite{neurips2023gaied_10_mcnichols,pankiewicz2023large,BEWERSDORFF2023100177,DBLP:journals/corr/abs-2307-02018,GPT4Hints-GPT3.5Val,pardos2023learning}, and facilitating the development of collaborative systems~\cite{2022.EDM-short-papers.54,neurips2023gaied_8_lee,neurips2023gaied_38_schmucker}. Few works have also used LLMs for learner modeling in programming domains~\cite{DBLP:journals/corr/abs-2310-10690} or for simulating students' behaviors as a basis for an interactive tool for teacher training~\cite{DBLP:conf/lats/MarkelOLP23}. However, despite their proficiency in linguistic tasks, LLMs often fall short in decision-making in a constrained environment, a domain where RL agents excel due to their inherent capability to make feasible decisions within a given environment~\cite{scienceworld}. 



Given the strengths and limitations of RL and LLM-based agents, recent works have investigated the integration of LLMs with RL to design agents that overcome the individual limitations of these agents. For instance, this integration has been used to substantially improve reward design and exploration efficiency in various domains~\cite{nottingham_embodied_2023,DBLP:conf/nips/LiPPDWF0HAAAM0Z22,du_guiding_2023,DBLP:conf/iclr/KwonXBS23}. However, most of these approaches have focused on the use of LLMs for training, bearing the risk of taking on LLMs' limitations in decision-making in constrained environments.
 
In this paper, we investigate the integration of RL and LLMs to create agents with enhanced generalizability in text-based educational environments, focusing on employing the LLM in the inference phase. To support our investigations, we present a novel text-based simulation benchmark, \pst, adapted from the PharmaSim virtual pharmacy environment designed for practicing diagnostic conversations. We present three types of agents: (i) RL-based agents employing natural language based representations, (ii) LLM-based agents invoked through prompting, and (iii) hybrid models where LLMs assist RL agents in the inference phase. 

We extensively evaluate all agents based on their ability to engage in effective diagnostic conversations and achieve accurate diagnoses on the \pst benchmark, focusing on their performance across a range of rephrased scenarios across diverse patient profiles. With our experiments, we aim to address three research questions: Which agent type demonstrates overall superior performance in conducting effective diagnostic conversations and achieving accurate diagnoses for all available patients \textbf{(RQ1)}? How does reflective prompting influence the diagnostic performance and conversation quality of LLM-involved agents \textbf{(RQ2)}? How do diagnostic performance and conversation quality vary among different agent types across diverse patients \textbf{(RQ3)}?
Our results demonstrate that a specific type of LLM-assisted RL agent outperforms all other agents in a combined score by effectively balancing accurate diagnosis along with high-quality diagnostic conversations. The source code and benchmark are released on GitHub.\footnote{https://github.com/epfl-ml4ed/PharmaSimText-LLM-RL \label{footnote.github}}
\section{Related Work}
Given our focus on integrating RL agents and LLMs to create generalizable learner models, we review prior work in developing learner models, explore the growing field of intelligent agents in text-based interactive games and finally discuss recent advancements in integrating RL and LLMs.

\noindent \textbf{Learner agents in educational environments.} There is a large body of research \cite{Kser2023} on simulating learners in online environments. Existing research provides rich, but not generalizable learner representations, for example by generating cognitive models from problem-solving demonstrations (e.g., SimStudent~\cite{DBLP:conf/edm/LiCKM11}) or simulates learners from student models in a data-driven way~\cite{Corbett2005KnowledgeTM,DBLP:conf/edm/FauconKD16,DBLP:conf/edm/BotelhoAH16}, leading to less rich, but more generalizable representations. RL is a promising tool to address these limitations. However, in the education domain, this framework has been primarily applied for pedagogical policy induction~\cite{DBLP:journals/tlt/WhitehillM18,pickthemom,zhou_hierarchical_2019,DBLP:journals/cogsci/RaffertyBGS16}, providing tailored hints~\cite{DBLP:conf/its/BarnesS08, DBLP:conf/edm/EfremovGS20}, generating educational content~\cite{DBLP:conf/nips/AhmedCEFGRS20,padurean2024neural}, and assessing interventions in educational platforms~\cite{rafferty_statistical_2019,mui2021multi}. Despite its potential, the exploration of RL-based learner agents for effective operation in learning environments remains limited~\cite{maclellan_learning_2021,DBLP:conf/iclr/BunelHDSK18}. Prior work has for example used Proximal Policy Optimization (PPO) for designing learner models in intelligent tutoring systems~\cite{maclellan_learning_2021} or employed neural and symbolic program synthesize to create student attempts in a block-based programming environment~\cite{DBLP:journals/corr/abs-2205-01265}. In this paper, we develop a series of learner agents for an open-ended educational environment.

\noindent \textbf{Agents for text-based interactive games.} The growing interest in developing intelligent agents for text-based interactive games, especially those that mimic real-world scenarios~\cite{scienceworld, Zhou2023SOTOPIAIE, Pan2023DoTR}, has led to diverse methodologies encompassing RL~\cite{DRRN}, behavior cloning (BC)~\cite{scienceworld}, and prompting LLMs~\cite{react,majumder_clin_2023}. A well-known example is the game ScienceWorld~\cite{scienceworld}, where players engage in scientific experiments through environment exploration and interaction. Within the RL framework, state-of-the art employs deep reinforced relevance networks (DRRNs)~\cite{DRRN}, treating text-based interactions as partially-observable Markov decision processes (POMDPs), and learning distinct text representations for observations and actions to estimate Q-values via a scorer network. Within the LLM domain, LLM-based strategies use prompts at each interaction step for strategic planning and action selection. While some studies~\cite{react} engage in a single interaction round with the environment, others~\cite{DBLP:journals/corr/abs-2303-11366/reflexion,majumder_clin_2023} use a multi-round approach, facilitating iterative refinement through repeated attempts. In this paper, we develop a series of agents for a text-based educational environment simulating real-world scenarios happening in a pharmacy.

\noindent \textbf{RL and LLM integration.} Recently, LLMs have been used to assist RL agents in various tasks, demonstrating notable advancements in reward design and exploration efficiency. For example, \cite{du_guiding_2023} utilized text corpora to pre-train agents, thereby shaping their exploration by suggesting goals based on the agents' current state descriptions. Furthermore, \cite{DBLP:conf/iclr/KwonXBS23} proposed a novel approach to simplify reward design by employing LLMs to generate reward signals from textual prompts that describe desired behaviors. 
In a similar vein, \cite{nottingham_embodied_2023} showcased the innovative application of few-shot LLM prompting to hypothesize world models for RL agents, which improves training sample efficiency and allows agents to correct LLM errors through interaction with the environment. While these studies highlight the synergistic potential of integrating LLMs with RL techniques to achieve more objective-aligned agent behaviors, directed exploration, and efficient training processes, the use of LLMs in the training phase bears the risk of carrying over their limitations in decision-making in constrained environments. A notable gap, therefore, remains in using LLMs to assist RL agents during the inference phase. Specifically, the current body of work has not addressed the use of LLMs to aid RL agents in adapting and transferring their learned skills to novel environments or tasks post-training. In our work, we aim to bridge this gap by focusing on utilizing LLMs as assistants for RL agents during generalization to new settings.


\section{PharmasimText benchmark}
We created \pst, a text-based interactive environment, as an infrastructure for developing language agents capable of handling text-based learning tasks and generalizing in them. 
\pst is an interactive text-based environment designed based on PharmaSim, a scenario-based learning platform. 
It simulates real-world interactions between a pharmacist and a patient in a pharmacy setting. 
This benchmark includes more than 500 scenario variations that can be used for developing and evaluating learner agents.

\begin{figure}[t]
    \centering
    \includegraphics[width=0.5\textwidth]{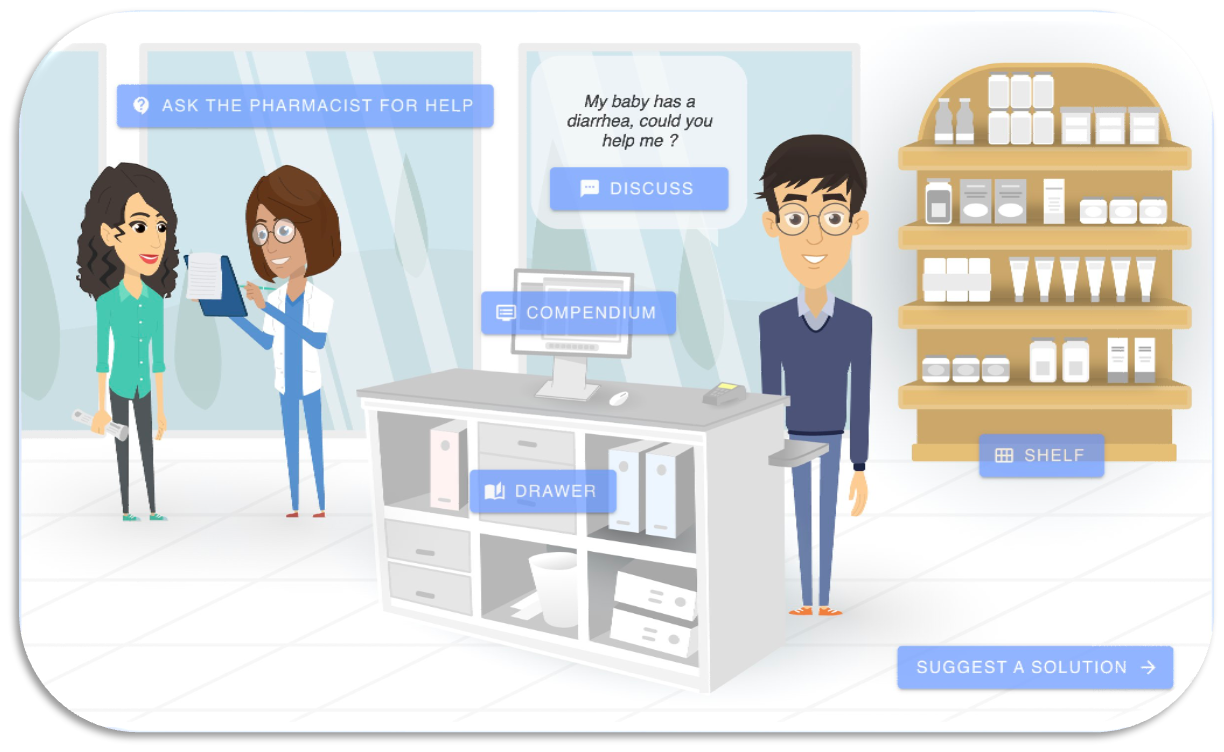}
    \caption{\textbf{'Father Inquiry' Scenario in PharmaSim} - A simulated pharmacy setting designed for practicing diagnostic conversational skills, where participants engage with a father seeking guidance for his infant child's diarrhea.}
    \label{fig:pharmasim}
    \Description{An overview of the Father Inquiry scenario in the PharmaSim environment which shows a father asking for help for his baby's diarhea}
\end{figure}

\begin{figure*}[ht]
    \centering
    \includegraphics[width=\textwidth]{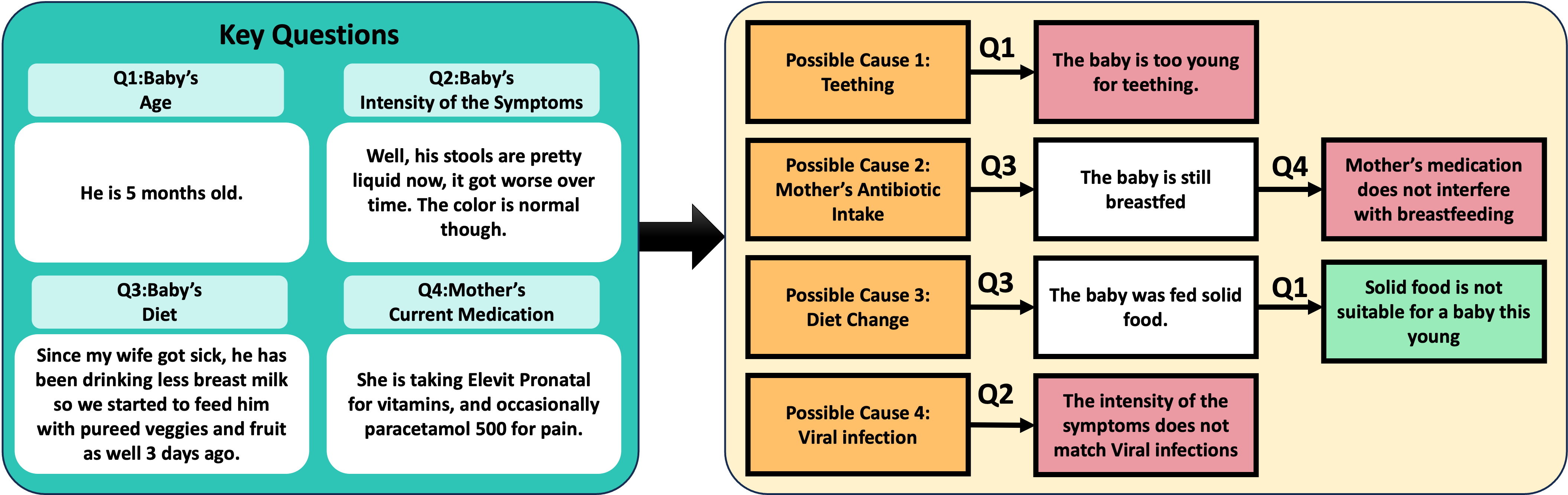}
    \caption{\textbf{Diagnostic Strategy in the 'Father Inquiry' Scenario of PharmaSim}, depicting the process of identifying the most likely cause of an infant's diarrhea. Players must pose four key questions to the father to collect crucial information, enabling the determination of the most probable cause of the child's diarrhea among four potential causes.}
    \label{fig:progress}
    \Description{Diagnostic Strategy in the 'Father Inquiry' Scenario of PharmaSim. There are 4 key questions including Baby's age, baby's intensity of the symptoms, baby's diet, and mother's current medication with the father's response to to them in the left. In the right, the four possible causes teething, mother's antibiotic intake, diet change, and viral infection are investigated by different key questions.}
\end{figure*}

\subsection{PharmaSim}
PharmaSim is a scenario-based learning environment designed to support the development of diagnostic skills. 
In each scenario, a patient comes to the pharmacy and asks for help with a specific problem. The player needs to identify different possible causes of this problem and mark how probable they are while interacting with the environment. Specifically, there are six different types of interactions: asking questions to the patient, seeking help from the pharmacist, searching about different kinds of medicine, looking for the specifications of products available on the shelf, reading about issues related to the problem, and offering a solution, which ends the game and moves the player to the post-test phase. In the post-test phase, players need to list three possible causes, rate their probability, and give an explanation for each of them. The determination of these likelihoods that leads to finding the most probable cause significantly depends on a set of patient inquiries containing essential information, which we henceforth refer to as \textit{key questions}.
%

Currently, two different scenarios designed with insights from human experts are available in the game. For example, in one scenario (see Fig. \ref{fig:pharmasim}), a father visits the pharmacy looking for help with his infant child's diarrhea. 
The scenario presents four probable causes for the child's condition.  The player is required to ask four \textit{key questions} to the father to gather the essential information needed to find the most probable cause behind the child's diarrhea. The relation between these \textit{key questions} and the most probable cause of the child's diarrhea is illustrated in Fig. \ref{fig:progress}. For instance, inquiring about the child's age enables the player to deduce that teething is an improbable cause due to the child's young age.

\subsection{PharmaSimText} 
To develop our benchmark, several modifications to PharmaSim were implemented.

\noindent \textbf{Migration to a text-based environment.}
As the first step, we did two adaptions to PharmaSim to migrate it to a text-based environment. First, we simplified interactions to two types of actions: asking questions to the patient about various characters phrased similar to PharmaSim as "\texttt{I want to know about the {character}'s {topic}.}" and advancing to the post-test by proposing a solution as "\texttt{I want to suggest a solution.}". Second, we modified the post-test questions to offer a feasible assessment for the agents. To this end, we revised the three causes question to focus solely on the most probable cause of the patient's issue. 


\textbf{Extension of available scenarios.} In the next step, we focused on enriching \pst and enhancing its complexity. For this purpose, we expanded the two scenarios available in the original environment across three dimensions: (1) introducing new patients, (2) varying the scenarios to alternate the most probable cause of each patient's problem, and (3) diversifying patient responses by rephrasing them. Given the scale of extension, relying solely on human expertise was impractical. Instead, we leveraged the generative capabilities of LLMs combined with human insights to develop the scenarios in \pst. Prior to prompting LLMs for creating scenarios, we structured our expanded scenarios to align with the pharmacy assistant training curriculum of Switzerland. We gathered a set of health problems from the curriculum, assigning each to a fictional patient with a specified age and gender. We further identified a range of illnesses from the curriculum's textbooks, known to manifest symptoms relevant to the chosen problems.

\textbf{Prompting LLMs for scenario creation.} The scenario creation process involved three steps: (1) we prompted the LLM to generate a list of \textit{key questions} aimed at diagnosing the most probable cause of the patient's problem, (2) the LLM was tasked to simulate patient responses, assuming each illness on the list was the most probable cause behind their problem, and (3) the LLM was prompted to generate answers to common patient inquiries done by pharmacists. We used GPT-4 as the LLM for scenario creation; the exact prompts employed can be found on our GitHub repository (link provided in Footnote~\ref{footnote.github}). To ensure realism and applicability, a human expert has reviewed all of the scenarios and provided feedback including minor changes which were reflected in the final version of the scenarios. Additionally, the LLM was employed to diversify existing patient responses through paraphrasing, enhancing the scenarios' complexity. To further augment this complexity, fictional characters were introduced as distractors, enabling players to engage in more nuanced interactions.

\newcolumntype{C}[1]{>{\centering\arraybackslash}p{#1}}
\newcolumntype{Y}{>{\centering\arraybackslash}X}
\begin{table*}[htbp]
\centering

\rowcolors{2}{gray!10}{gray!40}
\begin{tabularx}{\textwidth}{C{2cm} C{2.5cm} Y C{2.5cm}}
\toprule
\textbf{Problem} & \textbf{\# of Possible Causes} & \textbf{Possible Causes} & \textbf{\# of Key Questions} \\
\midrule
Infant Diarrhea & 4 & Change of diet, Teething, Current medication of the mother, Viral Infection & 4 \\
Breastfeeding-related & 6 & Engorgement, Plugged Ducts, Cracked Nipples, Mastitis, Thrush, Low Milk Supply & 7 \\
Urological & 4 & Prostate Hyperplasia, Cystitis, Urge Incontinence, Stress Incontinence & 6 \\
Skin-related & 10 & Sunburn, Insect Bites, Acne, Eczema, Athlete's Foot, Psoriasis, Rashes, Warts and Corns, Cold Sores, Neurodermatitis & 10 \\
Eye-related & 5 & Dry Eyes, Allergic Conjunctivitis, Pink Eye, Eye Strain, Stye & 11 \\
Gynecological & 8 & UTI, Cystitis, Kidney Stones, Overactive Bladder, Pregnancy, STI, Stress Incontinence, Fungal Infection & 8 \\
Joint Pain & 5 & Osteoarthritis, Muscle Sprains, Tendonitis, Bursitis, Gout & 9 \\
Sore Throat & 5 & Common Cold, Influenza, Sinusitis, Pharyngitis, Bronchitis & 7 \\
\bottomrule
\end{tabularx}
\caption{\textbf{Overview of \pst Scenarios.} Every task within the benchmark is centered on a unique health problem, which could stem from various causes. Players must ask several \textit{key questions} to arrive at a correct diagnosis.}
\label{PharmaSimText}
\end{table*}

\looseness-1\textbf{Statistics on the  PharmaSimText benchmark.} The obtained benchmark contains eight distinct scenarios, each revolving around a unique patient profile. Details about the patients can be found in Table \ref{PharmaSimText}. On average, each scenario presents seven potential causes for the patient's problem, resulting in a total of $47$ scenario variations. Patient responses in each variation are articulated in ten diverse phrasings to enhance the depth and variability. Furthermore, each scenario necessitates the identification of an average of $7.8$ \textit{key questions} by the player. As a result, \pst can provide an enriched environment for further studies on agents for text-based interactive tasks and agents' generalizability.

\begin{figure*}[t]
    \centering
        \includegraphics[width=\linewidth]{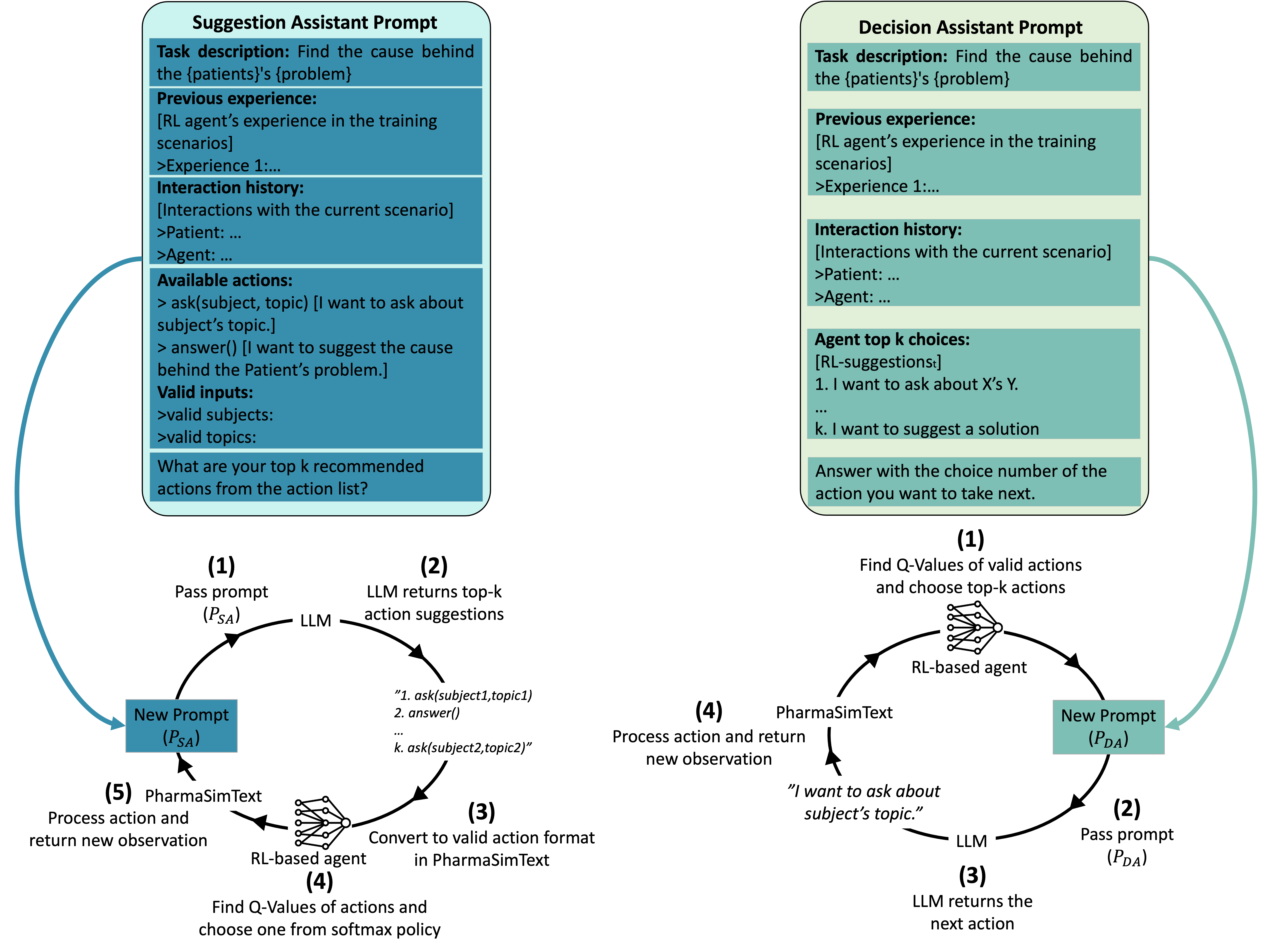}
    
    \caption{\looseness-1\textbf{LLM-assisted RL agents.} An LLM is prompted to assist the RL agent at the inference time to aid in generalization. In the Suggestion-Assisted RL (SA-RL) agent (left), the LLM suggests $k$ actions at each step for the RL agent to choose from. In the Decision-Assisted RL (DA-RL) agent (right), the LLM selects an action from the top-k choices provided by the RL agent.}
    \label{fig:assistance}
    \Description{The cycles LLM-assisted RL agents take to interact within the environment.}
\end{figure*}

\section{Agents for PharmaSimText}\label{agents}
We developed three types of agents for  \pst that embody various degrees of RL and LLM synergy: \rl agents, \llm agents, and \arl agents.

\subsection{RL-based Agents}\label{RL-based}
RL agents learn to interact within an environment by taking actions based on their current state and receiving feedback in the form of rewards or penalties for those actions~\cite{sutton}.
They try to maximize their obtained cumulative reward over time to effectively learn the best policy for achieving their goal within the environment.
One well-known method in RL involves estimating a metric called Q-value, which represents the expected future rewards for taking a certain action in a given state. 
Deep Q-Networks (DQNs)\cite{DBLP:journals/corr/MnihKSGAWR13/DQN} approximate these Q-values using deep neural networks, enabling handling of complex, high-dimensional environments by learning to predict the Q-values directly from the environmental states. 
DQNs are trained through interactions with the environment, using their experience to iteratively refine and make their estimations of Q-values more accurate.
 
Following previous work on text-based games, we utilized state-of-the-art, a DRRN \cite{DRRN} as the \rl agent for interacting with  \pst. The DRRN is designed to learn distinct representations for the text-based states and actions by employing two separate networks: the state encoder and the action encoder. A scorer network then evaluates these representations to estimate their Q-values. At a given step $t$ in the environment, the current state $s_t$ and the action taken $a_t$ are fed into the DRRN. Initially, $s_t$ and $a_t$ are encoded as sequences of word embeddings, which are subsequently processed by a Recurrent Neural Network (RNN) within both the state and action encoders to obtain respective embeddings for $s_t$ and $a_t$. Following the RNN layer, a Multi-Layer Perceptron (MLP) in each encoder refines these embeddings into more concise representations. These representations are then concatenated and fed into the scorer network's MLP, which yields an estimation of the Q-value $Q(s_t,a_t)$.

In our case, the valid actions at time step $t$ are interactions available in the environment presented to the agent as a list of sentences. After taking each action, the agent will receive a new observation $o_t$ that is formatted as: \texttt{Interaction type;} \texttt{Selected interaction;} \texttt{The patient's response}. For instance, in the scenario related to infant diarrhea if the agent decides to ask about the infant's age, the new observation will be formatted as: \texttt{Discuss;} \texttt{I want to know about the infant's age;} \texttt{He is 5 months old}. Therefore, the agent should consider the full history of its observations to comprehend its current state $s_t$ in the environment.
 
We introduced two modifications to adapt the original DRRN to our environment. First, we employed pre-trained sentence embeddings from fastText~\cite{bojanowski2016enriching} to generate text representations for both observations and actions. This choice was motivated by previous work showing that training the RNNs in the encoders of a DRRN with a loss function solely aligned with the RL objectives leads to unstable training and suboptimal embeddings~\cite{DBLP:conf/iclr/AmmanabroluH20/KGA2C}. Second, unlike the environments that DRRNs were proposed to tackle the tasks in, the observation at a given time step $t$ in \pst does not suffice for the agent to obtain a notion of the current state in the environment and the whole full observation history is needed as a part of context given to the agent. Therefore, we introduced a unit called the \textbf{state updater} before the state encoder that takes the previous embedded state $e(s_{t-1})$ and the new embedded observation $e(o_t)$ and returns the updated state after the current observation $s_t$. We experimented with five different methods in the state updater: mean pooling, max pooling, summation, an LSTM layer, and an LSTM layer with self attention. After a series of experiments, we observed the method based on summation led to the most stable training; therefore this method was adopted in our state updater. Formally, this method based on the summation of all the observation embeddings in the history, returns $e(s_t)=e(s_{t-1})+e(o_t)$ as the new embedded state $e(s_t)$.
 
\vspace{-0.2cm}

\subsection{LLM-based Agents}\label{LLM-based}
\looseness-1The agents based on LLMs prompt an LLM at each step of interacting with the environment to find the best next action to finish the task. These agents can either have only one trial or multiple trials to complete the task along with reflection on their strategy between each trial. We respectively denote these two agent types by \textit{none-reflective} and \textit{reflective}.
 
The \textit{none-reflective agent} interacts with the LLM by issuing a single prompt that contains the task description, the history of interactions (consisting of the agent's questions and the patient's responses), prior experience with the patient, and valid actions available at the current step to choose the most appropriate subsequent action. The task description is structured as \texttt{Find the cause behind the {patient}'s {problem}}, while the interaction history is presented as a dialogue between the patient and the agent, with action texts labeled as agent's questions and environment's feedback text as patient responses. To format the valid actions, each action type is formatted as a function along with its permissible input values, which the LLM can interpret. This is complemented by a descriptive text explaining the action's purpose. For instance, the interaction "I want to ask about the subject's topic" is formatted as \texttt{ask(subject, topic): Asking a question about the subject related to the topic}, followed by a list of valid subjects and topics. This meticulous formatting strategy plays an essential role in minimizing the likelihood of the LLM suggesting invalid actions.
 
Despite efforts to format valid actions to guide the LLM, there are instances where the LLM still proposes an action that is invalid within the \pst environment. In such cases, we implemented a strategy where the LLM was prompted to suggest an alternative action, repeating this process for a maximum of $k=3$ attempts. Should all suggested actions remain invalid, we selected the valid action that has the smallest distance in the natural language embedding space to the $k$-th suggested action. This approach ensures that the LLM's output is effectively grounded in the set of actions that are feasible within the environment.
 
The \textit{reflective agent} employs a prompting strategy akin to that of the \textit{none-reflective agent} to determine the optimal subsequent action. The \textit{none-reflective agent} prompt is augmented with a segment including learnings from prior engagements with the same patient having the same cause. This reflective process involves prompting the LLM to evaluate its previous strategies based on the observed outcomes after completing each trial. The agent then updates its textual memory of previous learnings, and the updated memory is used for prompting in the next trial. This approach was inspired by research on self-reflective LLMs, notably the continually learning language agent CLIN\cite{majumder_clin_2023}. Similar to CLIN, we constructed the learning memory using causal formats such as “X is necessary for Y” to guide future interactions. This mechanism enables the reflective agent to dynamically adapt and refine its approach, enhancing its decision-making process over time.

\subsection{LLM-assisted RL Agents}
The perspective of \rl agents remains limited to their experience during training, potentially hindering the performance in tasks with unfamiliar elements not encountered during their training. To address this, we leveraged LLMs' commonsense reasoning capabilities to augment RL agents' decision-making processes. As shown in Fig. \ref{fig:assistance}, we explored two methods for integrating LLM assistance: \textit{\saf (\sa)} and \textit{\daf (\da)}.
 
In the \sa approach, at a given time step $t$, the LLM is prompted to suggest a list of $k$ best actions to be taken at that state called $\text{LLM-Suggested}_t$. The actions' Q-values in $\text{LLM-Suggested}_t$ are then calculated by the RL agent, and the next action is sampled from the probability distribution obtained by taking softmax over the estimated Q-values. The prompting format here is similar to the \llm agents discussed in Section~\ref{LLM-based} containing the task description, the history of interactions, prior experience with the patient, and valid actions at that step. We set $k=5$ in the interaction steps and $k=2$ in the posttest steps.
 
In the \da approach, at a given time step $t$, we collect a list of $k$ most probable actions under the RL agent's policy $\text{RL-Suggested}_t$. Then, an LLM is prompted to choose the best action among the actions in $\text{RL-Suggested}_t$. The prompting used for this task contains the task description, the history of interactions, prior experience with the patient, and the actions in $\text{RL-Suggested}_t$. Therefore, the LLM acts as a decision assistant for the RL agent.
Notably, in our implementation, we set $k=5$ in the interaction steps and $k=2$ in the post-test steps.
 
Based on whether the LLM is given an opportunity to reflect on its past decisions or not, we obtain two versions of \da and \sa approaches, which we distinguish via reflective/none-reflective prefixes. Thus, we study four \arl agents: \textit{none-reflective-\da}, \textit{reflective-\da}, \textit{none-reflective-\sa}, and \textit{reflective-\sa}. 
\section{Experimental Evaluation}
\begin{figure}[t]
\centering
\includegraphics[width=\linewidth]{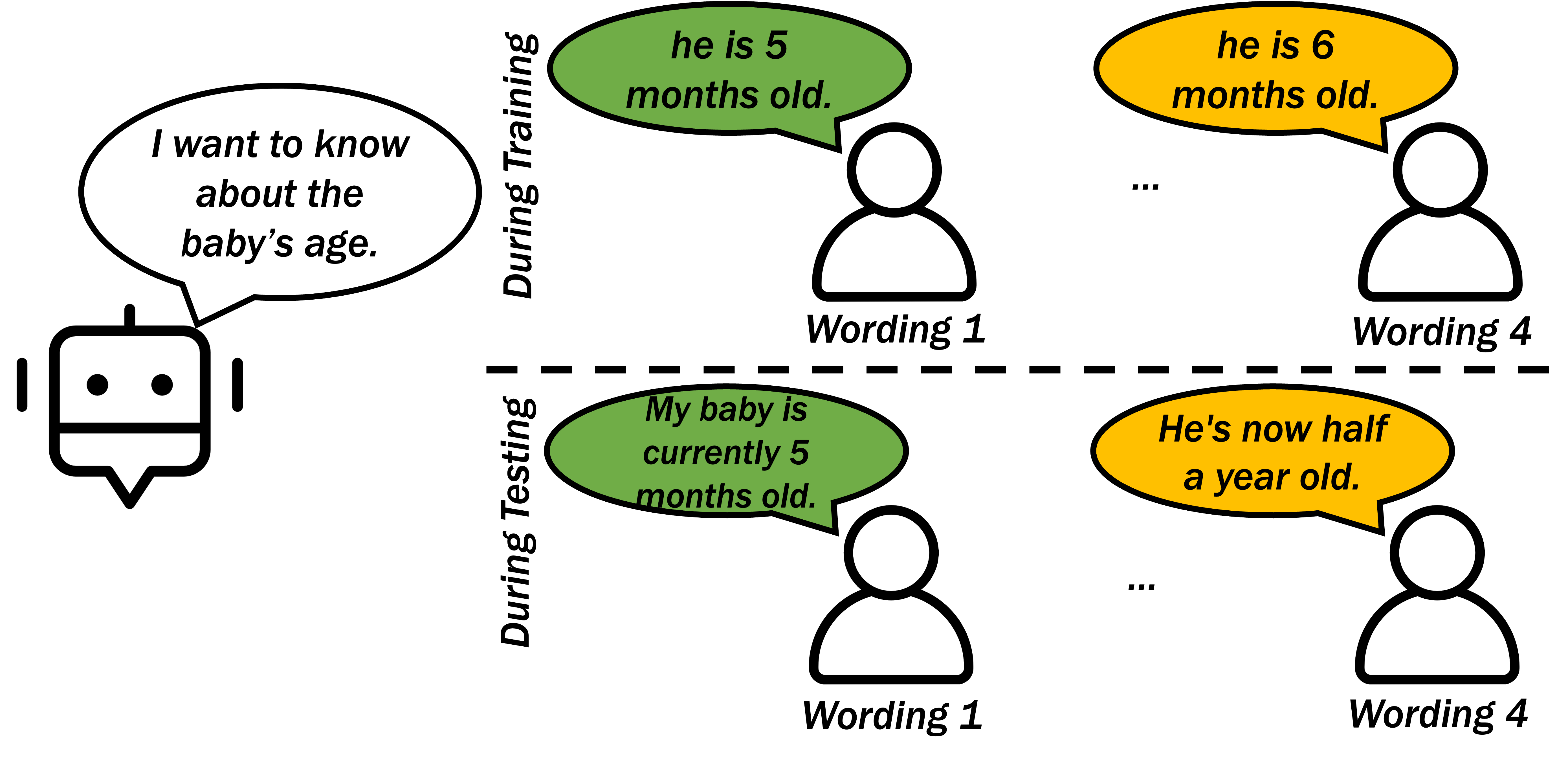}
\caption{\textbf{Generalization task}, requiring the agents to generalize over different wordings of a scenario.}
\label{fig:wording}
\Description{an example of rephrased wordings for the question about baby's age in which he is 5 months old is rephrased to my baby is currently 5 months old}
\end{figure}
\begin{figure*}[t]
        \centering
        \includegraphics[width=\linewidth]{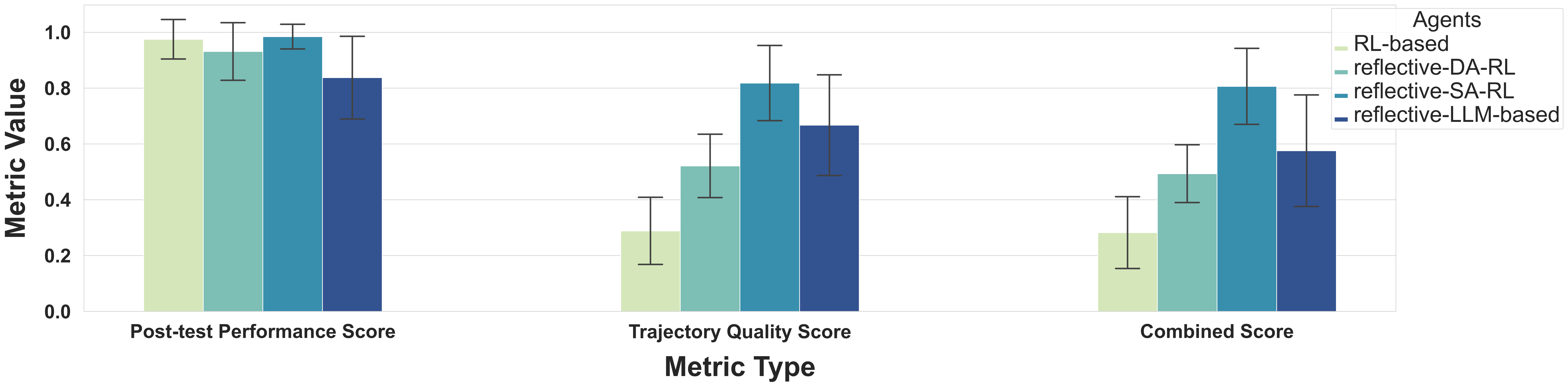}
    \caption{\textbf{Agent Performance on PharmaSimText.} \perf (left), \traj (middle), and \comb (right) of the \rl agent, the reflective-\da agent, the reflective-\sa agent, and the reflective-
\llm agent. In the \sa agent, the LLM suggests $k$ actions at each step for the RL agent to choose from. In the \da agent, the LLM selects an action from the top-k choices provided by the RL agent. Scores are averaged across all patients in \pst.}
    \label{fig:wording_result}
    \Description{Agent Performance on PharmaSimText}
\end{figure*}
We evaluated our agents in \pst to assess which agent type demonstrates the most effective diagnostic conversations and accurate diagnoses among all patients \textbf{(RQ1)}, to investigate the impact of reflective prompting on the diagnostic performance and interaction quality of LLM-involved agents \textbf{(RQ2)}, and to explore how diagnostic performance and conversation quality vary among the different agent types when confronted with different patients \textbf{(RQ3)}.

\subsection{Experimental Setup}
Our evaluation was focused on the generalization capabilities of the agents, specifically their ability to navigate tasks featuring not previously encountered elements. We assessed the agents' generalizability across rephrased versions of already-encountered scenarios, aiming to measure their reliance on the precise wording of these scenarios. Figure \ref{fig:wording} provides insight into our evaluation methodology for generalization, illustrating the diversity created by rephrased answer options in a specific scenario.

We defined agent success in a subtask based on two aspects: identifying the most probable cause of the patient's problem and asking the \textit{key questions} in the conversation. Here a subtask denotes the combination of a cause and a wording. We therefore introduced three metrics: 
\vspace{-2mm}
\begin{itemize}[parsep=2.5pt]
    \item \perf: binary indicator of correct diagnosis of the patient’s problem. It measures the agent's ability to identify the most probable cause of the patient's problem.
    \item \traj: fraction of key questions involved in the agent’s conversation. It measures the quality of the agent's conversation.
    \item \comb: product of the \perf and \traj. It measures both the above elements together.
\end{itemize}
\vspace{-2mm}

\subsection{Agent Training and Evaluation}
We developed and trained all of the agents separately for each patient. In this process, different wordings of subtasks leading to the same cause were split randomly to a training, validation, and test set. Therefore, the training, validation, and test sets included subtasks of all of the causes available for a patient in distinct wordings. Specifically, the agents saw all the causes during training and validation, but not all wordings. In our experiments, $80\%$ of the available wordings for each cause were used for training and the remaining wordings were split in half for the validation and test set. 

The \rl agents were trained using subtasks from the designated training set being given a random subtask at each episode of interaction with the environment. At a given time step $t$, the agent took an action sampled from a softmax policy obtained from the Q-values of all of the actions available. The randomness of the softmax policy was controlled using a temperature decaying from 1 to 0.001 linearly during the training. After each interaction, the agent was rewarded using a reward function that awarded the agent a positive reward of +1 when it successfully completed the posttest and penalizes with -1 otherwise. Moreover, each interaction of the agent was penalized by a small negative reward of -0.01. 

Following each iteration of training, these agents underwent an evaluation phase using subtasks from the validation set. The iteration that yielded the highest average \perf on the subtasks in validation set was used for testing and also served as the foundation for the RL component within the \arl agents. 

The agents that had an LLM involved in their structures used the GPT-4 model. The \llm agents initially gain experience through interactions within the training subtasks. This acquired experience is subsequently leveraged during their engagement with the test subtasks.

\subsection{RQ1: Efficacy of Different Agent Types}
In our first analysis, we aimed to assess the agents' efficacy in diagnostic dialogues and accuracy in diagnoses aggregated over all patients. Figure \ref{fig:wording_result} illustrates the  \perf, \traj, and \comb of the different agents.

\begin{figure*}[t]
    \centering
    \includegraphics[width=1.0\linewidth]{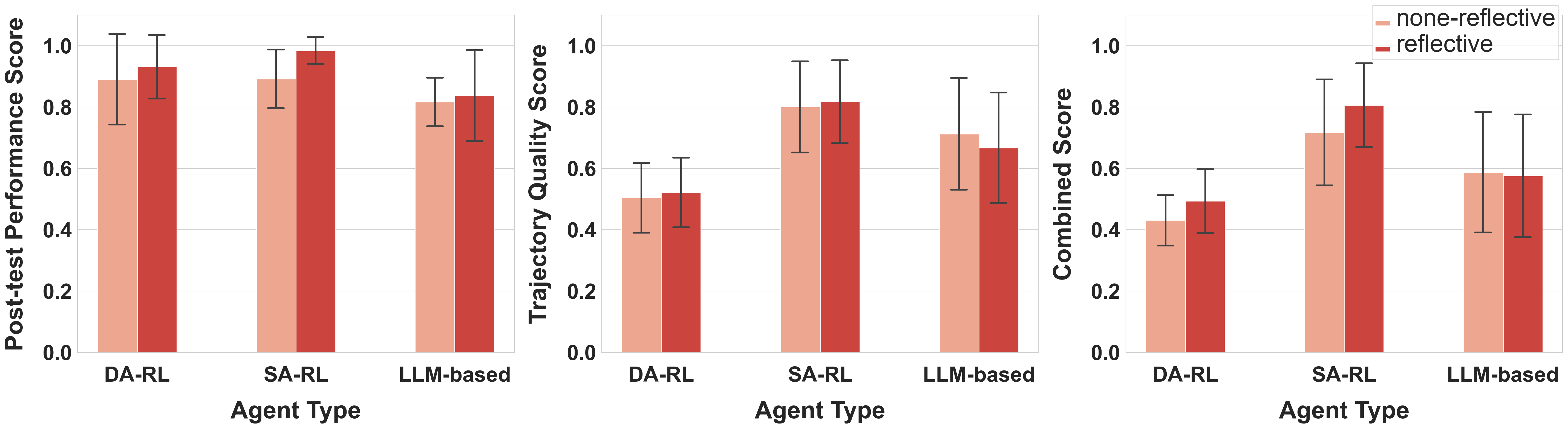}
    \caption{\textbf{Performance of reflective and none-reflective agents on PharmaSimText.} \perf (left), \traj (middle), and \comb (right) for none-reflective and reflective \da, \sa, and \llm agents. }
    \label{fig:reflective}
    \Description{Performance of reflective and none-reflective agents on PharmaSimText}
\end{figure*}

We observed that the \rl agent achieved a high \perf, indicating its ability to arrive at the correct diagnosis through a process of trial and error. However, this agent's approach often lacked the depth and nuance of a meaningful diagnostic conversation, reflected in its low \traj. This observation is probably due to its lack of background knowledge and common sense reasoning. Conversely, the \llm agent exhibited a superior capacity for engaging in meaningful diagnostic dialogues, reflected in a higher \traj. However, the \llm agent exhibited a lower \perf than the \rl agent, indicating that its ability to consistently reach the correct diagnosis is inferior compared to the \rl agent.

In examining the \arl agents, both \da and \sa agents surpassed the \llm agent in \perf, indicating that integrating LLM with RL generally improves diagnostic precision of purely \llm agents. Notably, the \sa agent exhibited superior \perf closely mirroring that of the \rl agent. The \da's relative under-performance may have stemmed from its longer trajectories compared to the \rl agent, leading to unfamiliar states where the DRRN struggled to provide accurate diagnoses, thereby affecting the \da's RL-driven suggestions. Furthermore, in terms of engaging in quality diagnostic dialogues, the \sa agent was also superior to the \da agent. This superiority is likely due to the RL framework's preference for shorter, more direct solutions, which reduced the action quality suggested by the DRRN in prolonged interactions. This effect was more pronounced in the \da agent, potentially constraining the quality of diagnostic conversations.

In the comparison of the agents in the \comb, the \sa agent emerged as the standout performer. Unlike its counterparts, the \sa agent adeptly navigated the dual challenges posed by the benchmark, demonstrating both a high conversation quality and diagnostic accuracy. This achievement highlights the \sa agent's unique capacity to capture the strengths of both \rl and \llm agents through the addition of suggestion-based assistance from LLMs to the RL agents' decision-making process.

To further investigate the results, we performed additional statistical tests. A Kruskal-Wallis test shows significant differences between the agents for the \traj and \comb $(p_{trajectory}<0.0001\text{ and }p_{combined}<0.001)$ and a trend to significance for the \perf $(p_{performance}=0.052)$. Post-hoc comparisons using Mann-Whitney U tests with a Benjamini-Hochberg correction for the \comb indicate significant differences between 5 out of 6 pairs of agents supporting prior findings. For instance, the comparison between \rl agent and \sa agent resulted in a p-value smaller than 0.01, and for the comparison between \sa agent and \llm agent the p-value was smaller than 0.05.

\textit{In summary, the experimental outcomes highlight distinct strengths and weaknesses among the agents. The \rl agent demonstrated proficiency in achieving a high \perf score, but was hindered in engaging in effective diagnostic dialogues due to limited background knowledge. Conversely, the \llm agent excelled in conducting high-quality conversations by leveraging its extensive knowledge base, though with less accuracy in diagnoses. The hybrid \arl agents, \da and \sa, outperformed the \llm agent in diagnostic precision and surpassed the \rl agent in dialogue quality. The \sa agent achieved both a high conversation quality and diagnostic accuracy, illustrating its effective integration of LLM  and RL capabilities.}

\begin{figure*}[h]
    \centering    
    \includegraphics[width=\linewidth]{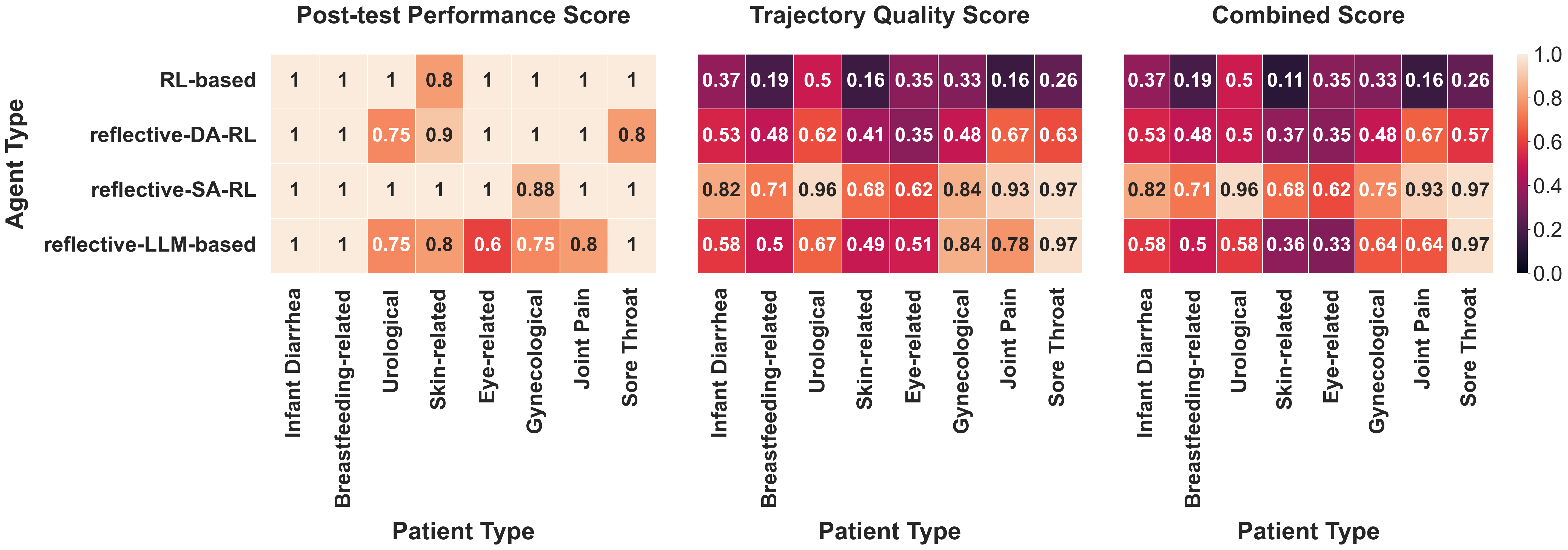}
    \caption{\textbf{Performance of different agents in interaction with different patients.} \perf (left), \traj (middle), and \comb (right) for \rl and reflective \sa, \da, and \llm agents.}
\label{fig:agents_vs_patients}
\Description{Performance of different agents in interaction with different patients}
\end{figure*}

\subsection{RQ2: Effect of Reflective Prompting}
In our second analysis, we aimed to explore the impact of reflective prompting on the efficacy of LLM-involved agents. As described in Section~\ref{agents}, none-reflective agents were limited to a single attempt, whereas reflective agents were given three attempts per subtask with opportunities for reflection. Figure \ref{fig:reflective} illustrates the \perf, \traj, and \comb for none-reflective and reflective \arl and \llm agents.
%

We observed a nuanced impact of reflective prompting on agent performance. Specifically, reflective prompting did not significantly impact the \comb of the purely \llm agent. For this agent, reflection led to shorter diagnostic conversations by eliminating what the agent considered redundant questions. However, this streamlining resulted in poorer conversation quality without significantly improving diagnosis accuracy, negating the potential diagnosis accuracy gains from reflection.
 
In contrast, the reflective process considerably enhanced the performance of the hybrid \arl agents. This improvement can be attributed to the reflective phase allowing the agents to reassess and refine their decision-making processes, leading to more accurate diagnoses. The performance boost was particularly notable in \sa agents, most likely due to their reliance on the LLM for suggesting potential actions during the interaction phase. This reliance provided a broader scope for reflection to influence decision-making, unlike \da agents where decisions were more heavily influenced by the \rl agent. This finding underscores the value of incorporating reflective mechanisms in enhancing the capabilities of hybrid agents.
 
\textit{In summary, our experiment revealed that reflective prompting has a different effect on \llm and \arl agents. For the \llm agents, reflective prompting led to shorter and lower quality diagnostic conversations, with no significant improvement in diagnostic accuracy. On the other hand, the \arl agents benefited from reflection, showing improvements in diagnostic accuracy. This enhancement was more pronounced for \sa agents, which rely more on LLM suggestions.}

\begin{figure}[t]
    \centering
    \includegraphics[width=\linewidth]{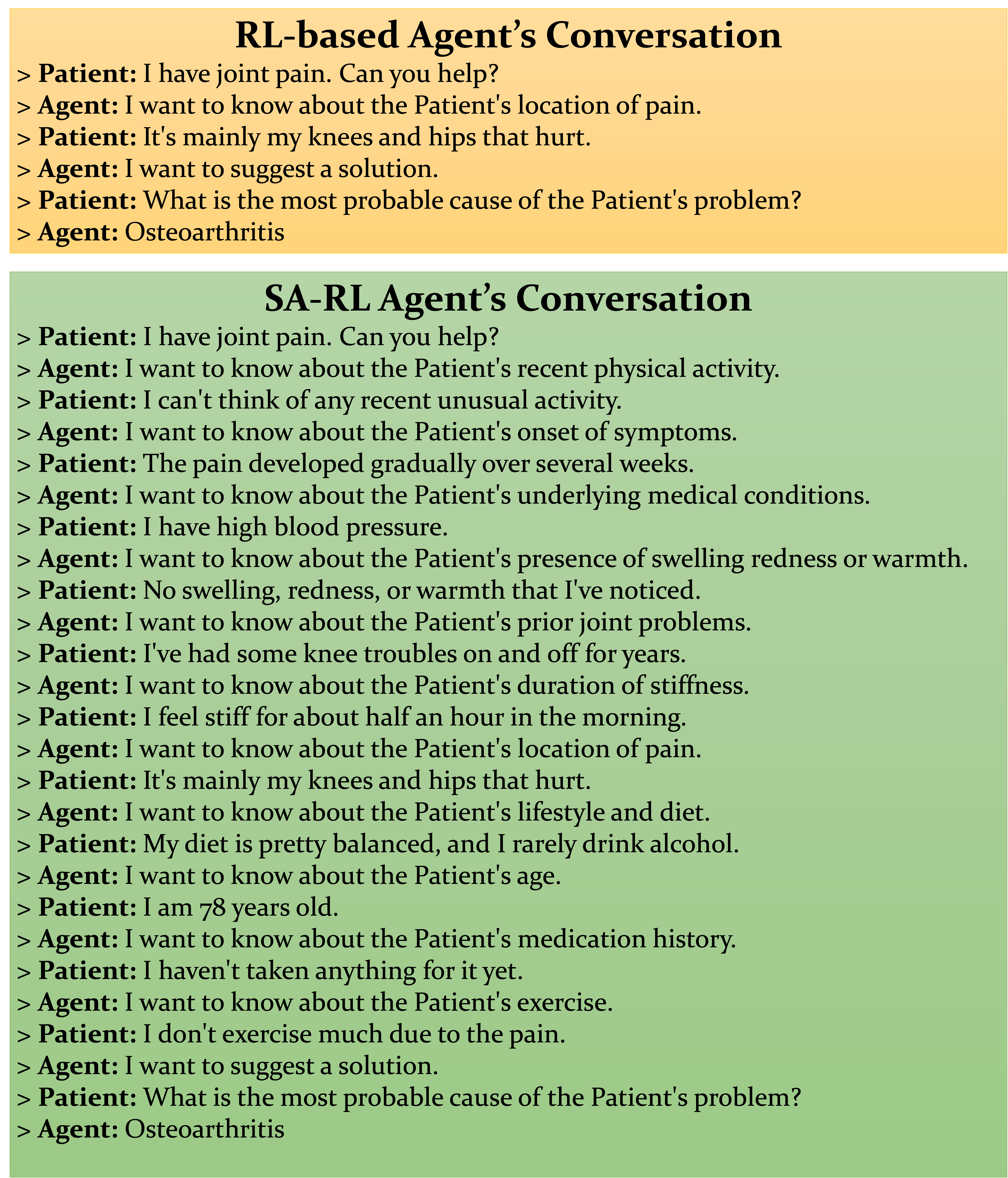}
    \caption{\textbf{Example diagnostic conversations} conducted by the \rl (top) and \sa agents (bottom) with the patient with joint pains in a test subtask with Osteoarthritis as the most probable cause.}
    \label{fig:conversation}
    \Description{Example diagnostic conversations with the patient with joint pains in a test subtask with Osteoarthritis as the most probable cause.
    RL-based Agent’s Conversation and
SA-RL Agent’s Conversation
}
\end{figure}

\subsection{RQ3: Agent Efficacy for Different Patients}
In our final analysis, we investigated the performance of our agents across the different patients. Figure \ref{fig:agents_vs_patients} illustrates the \perf, \traj, and \comb for each patient averaged over all of the subtasks available for that patient in \pst for the \rl agent as well as the reflective \sa, \da, and \llm agents. 

We again observed that the \rl agent showed superior \perf across all patients, while the \llm agent was not able to identify all causes correctly for five out of the nine patients. The \arl agents managed to overcome this limitation, with the \sa agent showing superior performance than the \da agent. The opposite result was found for the \traj. While the \llm agents conducted high-quality diagnostic dialogues, the \rl agent exhibited a suboptimal \traj for all of the patients, often incorporating merely one or two key questions within its diagnostic conversations, highlighting the extent of its deviation from an effective diagnostic interaction. Again, the \arl agents overcame this limitation, with the \sa agent generally showing the highest \traj scores.
 
Our examination of the \comb revealed that, except for the \sa agent, most agents encounter difficulties in scenarios related to Skin and Eye conditions. A closer inspection of their \perf and \traj metrics suggested that these agents face challenges in different facets of the scenarios related to these specific patients. A particularly noteworthy observation is the superior performance of the \sa agent, which overcomes the limitations of purely \rl and \llm agents across all patient categories.

\looseness=-1 Given the inferior performance of the \rl agent in the \traj, we examined the dialogues generated by the \rl agent and the \sa agent within an identical scenario that resulted in a correct diagnosis, as illustrated in Fig. \ref{fig:conversation}. This comparison reveals a pronounced contrast in the conversational dynamics of these two agents. The dialogue led by the \sa agent exhibits a flow that is markedly more reminiscent of human-like interaction, in contrast to the \rl agent's brief conversation. Notably, the \rl agent's approach is characterized by posing a single key question before directly drawing a conclusion. In comparison, the \sa agent engages in a more thorough inquiry, covering a broader spectrum of key questions in a logically sequential manner.

\textit{In summary, the hybrid \arl agents manage to ovecome the limitations of solely \rl and \llm agents, with the \sa agent demonstrating superior performance across all patients. The \rl agent exhibits a behavior characterized by short conversation, limiting interactions to very few key questions, while the \sa agent follows a more human-like behavior.}

\vspace{-0.2cm}
\section{Discussion and Conclusion}
In this paper, we explored integration of RL and LLMs to enhance learner models in educational technologies. While RL-based agents show promise in structured learning tasks, they struggle with open-ended environments and skill generalization. Conversely, LLMs excel in generating student-like responses, but fail in constrained action spaces. By combining RL and LLMs, we aimed to develop more generalizable agents for text-based educational settings. We assessed our agents, including \rl, \llm, and hybrid models, on their ability to conduct diagnostic conversations and make accurate diagnoses in our novel benchmark \pst.

Specifically, we were interested in answering the following three research questions: Which agent type demonstrates overall superior performance in conducting effective diagnostic conversations and achieving accurate diagnoses for all available patients \textbf{(RQ1)}? How does reflective prompting influence the diagnostic performance and conversation quality of LLM-involved agents \textbf{(RQ2)}? How do diagnostic performance and conversation quality vary among different agent types across diverse patients \textbf{(RQ3)}?

To address our first research question, we assessed four agents: one RL-based, one LLM-based, and two integrating LLMs with RL, in rephrased versions of the scenarios related to different patients in \pst that the agents had not seen before. Effective diagnostic conversations require high-quality conversations and accurate diagnoses. The RL agent excelled in finding the correct diagnosis but struggled in comprehensive diagnostic dialogues due to its limited knowledge. The LLM agent was adept in high-quality diagnostic conversations but tended to misdiagnose patients. LLM-RL integrations were able to address these limitations by enhancing the diagnostic accuracy compared to the \llm agent and the conversation quality compared to the \rl agent. Among all agents, the \sa agent achieved the best combination of diagnostic accuracy and conversation quality.
 
The second research question investigated the benefits of reflective prompting of the LLMs in the LLM-involved agents. To answer this question, we compared the reflective versions of three LLM-involved agents with their none-reflective counterparts. In prior works, reflection showed noticeable improvements in task completion of prompted LLMs \cite{DBLP:journals/corr/abs-2303-11366/reflexion,majumder_clin_2023}. Therefore, we hypothesized a noticeable drop in the performance of the LLM-involved agents after confining them to only one trial. Our results showed a mixed effect for reflection in the solely LLM-based agent and the hybrid agents. For the LLM-based agent, the reflection improved the diagnostic accuracy of the agent, but it decreased the quality of the agent's conversation by shortening its trajectory. For the hybrid agents, the reflective process increased the diagnostic accuracy. We therefore conclude that the effect of reflective prompting depends on the agent type.
 
To address the third research question, we analyzed the agents over the three metrics for each of the patients separately. We observed that the agents did not struggle with similar patients. In our subsequent analysis, we looked at an example of the conversations done by the \rl agent and the \sa agent, and we observed that while the \rl agent conversation seemed rushed, the \sa's conversation seemed human-like and followed a sequential logic.

One of the limitations of this work is the focus on generalization at a single level of rephrased versions of the scenarios. A few possible generalization levels available \pst are: generalizing to a new wording of a known scenario (wording generalization), to a new diagnosis of a known patient (subtask generalization), and to a new patient (task generalization). Our presented experiments are limited to the wording generalization. Further research should be done within different generalization levels to evaluate current agents and propose new agent frameworks that consider the models’ confidence in integration and leverage LLM insights for rapid adaptation of \rl agents to new tasks. Moreover, our proposed reflective process showed limitations in improving the LLM-based agents. This suggests a need for further research for improved reflection in the interactive format of the \pst benchmark. Moreover, future research should consider evaluating the similarity of behavior of these agents to human students to further facilitate their use cases such as evaluating learning environments and collaborative learning.
 
To conclude, the proposed LLM integration approach represents a promising step towards agents with generalization capabilities in open-ended text-based educational environments. Furthermore, our implemented benchmark facilitates further research in developing agents with generalization capabilities at a higher level.

\section{Acknowledgements}
We thank Dr. Jibril Frej and Dr. Ethan Prihar for their expertise and support. This project was substantially financed by the Swiss State Secretariat for Education, Research and
Innovation (SERI).

%
\balance
\bibliographystyle{unsrt}
\bibliography{references}
%
\balancecolumns
\end{document}